\documentclass[10pt, a4paper]{article}

\usepackage[final]{lrec2026}
\usepackage{booktabs}
\usepackage{todonotes}
\usepackage[dvipsnames]{xcolor}
\usepackage{amssymb}
\usepackage{amsmath}
\usepackage{cleveref}
\usepackage{enumitem}
\usepackage{multirow}
\title{SommBench: Assessing Sommelier Expertise of Language Models}

\name{
William Brach$^{1,2}$, Tomas Bedej$^{2}$, Jacob Nielsen$^{3}$, Jacob Pichna$^{2}$, \\
\textbf{\large Juraj Bedej$^{2}$, Eemeli Saarensilta$^{2}$, Julie Dupouy$^{2}$,} \\
\textbf{\large Gianluca Barmina$^{3}$, Andrea Blasi Núñez$^{3}$,} \\
\textbf{\large Peter Schneider-Kamp$^{3}$, Kristian Košťál$^{1}$, Michal Ries$^{1}$, Lukas Galke Poech$^{3}$}
}

\address{
$^{1}$Slovak University of Technology, Bratislava, Slovakia \\
$^{2}$sommify, Helsinki, Finland \\
$^{3}$University of Southern Denmark, Odense, Denmark \\
\\
\textit{Correspondence : william.brach@stuba.sk}
}

\abstract{
With the rapid advances of large language models, it becomes increasingly important to systematically evaluate their multilingual and multicultural capabilities. Previous cultural evaluation benchmarks focus mainly on basic cultural knowledge that can be encoded in linguistic form. Here, we propose SommBench, a multilingual benchmark to assess sommelier expertise, a domain deeply grounded in the senses of smell and taste. While language models learn about sensory properties exclusively through textual descriptions, SommBench tests whether this textual grounding is sufficient to emulate expert-level sensory judgment. SommBench comprises three main tasks: Wine Theory Question Answering (WTQA), Wine Feature Completion (WFC), and Food-Wine Pairing (FWP). SommBench is available in multiple languages: English, Slovak, Swedish, Finnish, German, Danish, Italian, and Spanish. This helps separate a language model's wine expertise from its language skills. The benchmark datasets were developed in close collaboration with a professional sommelier and native speakers of the respective languages, resulting in 1,024 questions for wine theory question answering, 1,000 examples for wine feature completion, and 1,000 examples of food-wine pairing. We provide results for the most popular language models, including closed-weights models such as Gemini 2.5, and open-weights models, such as GPT-OSS and Qwen 3. Our results show that the most capable models perform well on wine theory question answering (up to 97\% correct with a closed-weights model), yet feature completion (peaking at 65\%) and food-wine pairing show (MCC ranging between 0 and 0.39) turn out to be more challenging. These results position SommBench as an interesting and challenging benchmark for evaluating the sommelier expertise of language models. The benchmark is publicly available at \url{https://github.com/sommify/sommbench}.
 \\ \newline  
}

\begin{document}

\maketitleabstract

\section{Introduction}
Large language models achieve increasingly strong performance on multilingual benchmarks \cite{pomerenke2025ailanguageproficiencymonitor, hendrycks2021measuringmassivemultitasklanguage}, but a key open question concerns whether language models display consistent competencies in culturally-grounded, expert-level knowledge across languages, or whether they exhibit language-dependent behavior that reflects the cultural contexts embedded in their training data. This question matters as LLMs are deployed globally, where users interact in multiple languages and expect coherent, reliable answers \cite{ni2025surveylargelanguagemodel}.

\begin{figure}[!ht]
    \centering
    \includegraphics[width=\columnwidth]{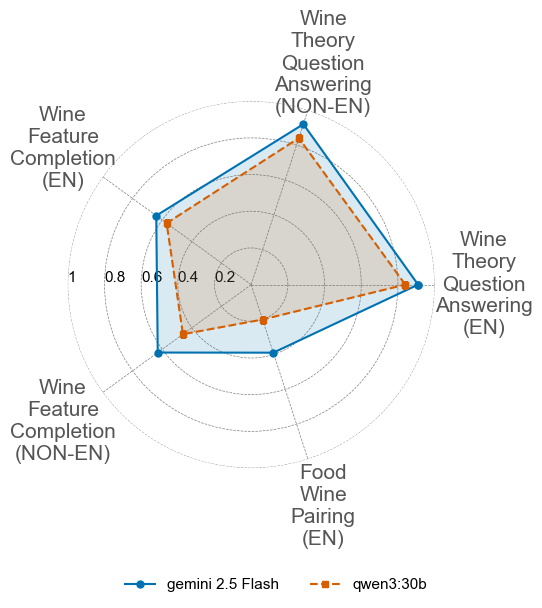}
    \caption{Performance of leading open and closed-source language models on the SommBench benchmark. The radar chart shows model accuracy (the higher, the better) in SommBench tasks revealing key differences in competencies between tasks and models.}
    \label{fig:figure_1}
\end{figure}

We investigate this question within the domain of sommelier expertise, a field rich with sensory, factual, and cultural knowledge~\cite{chen2022wineinformatics,eschevins2019expert,manske2005understanding}. 
Covering eight languages, our work examines whether a model's understanding of wine production methods, regional classifications, food pairings, or varietal properties changes depending on the language used in the prompt, i.e., when a model is asked in German or in Spanish about the characteristics of a Grüner Veltliner, does it provide consistent information, or does uneven training data across languages lead to contradictory responses? 

While some tasks like food-wine pairing are inherently culturally situated, the foundational knowledge underlying them should remain consistent across  languages~\citep{liu2025culturally,weeber2025political}. This makes sommelier expertise an interesting testbed for language models, posing two key challenges: interpreting subjective sensory properties learned purely from text, and maintaining cross-lingual consistency for factual knowledge.

To tackle these questions, we introduce SommBench, a multilingual benchmark covering 8 languages (\textit{English, Slovak, Swedish, Finnish, German, Danish, Italian, and Spanish}) that evaluates both cultural knowledge and cross-linguistic consistency through three complementary tasks: 

\begin{description}
    \item[Wine Theory Question Answering (WTQA)] 
A language model needs to answer multiple-choice questions, for which we use factual questions from established sommelier exams to test consistent knowledge recall across languages (as detailed in \S\ref{wtqa});

\item[Wine Feature Completion (WFC)] A language model needs to complete missing properties of a given wine, requiring cross-lingual prediction of sensory descriptors to test coherent representations (\S\ref{wfc}). 

\item[Food-Wine Pairing (FWP)] evaluating culturally-adaptive rationales by recommending pairings for culturally-specific dishes (\S\ref{fwp}). 
\end{description}
This structure allows us to distinguish between problematic inconsistency (e.g., contradictory facts about the same wine) and desirable cultural adaptation (e.g., culture-specific pairing suggestions).

Our experimental results with the most popular open and closed weight models on SommBench confirm that SommBench is an interesting and challenging benchmark. Even commercial frontier language models peak only at 0.65 aggregated score across tasks. Open weight models, on the other hand, peak at 0.52. Figure~\ref{fig:figure_1} shows exemplary results of best tested closed weights and open weights models.


Our main contributions are as follows:

\begin{itemize} 
\item We introduce SommBench, a novel benchmark in eight languages to evaluate LLMs on culturally grounded wine knowledge, totaling 3024 examples across three different tasks.
\item We demonstrate that, while leading models possess strong factual knowledge, they struggle with food \& wine pairing. In contrast, open-weights models display a performance decrease in non-English languages. 
\item Our analysis of food \& wine pairing tasks reveals that many models exhibit a common positivity bias, favoring approval over rejection..
\item We provide a comprehensive analysis of numerous closed and open-weights models, establishing baselines for cross-lingual consistency and culturally-aware competencies in the specialized domain of sommelier expertise.
\end{itemize}

\section{Related work}

The evaluation of LLMs is shifting towards specialized benchmarks. While multilingual datasets like Global-MMLU~\cite{singh2024global} test translated general knowledge, cultural benchmarks have begun to address culturally-situated evaluation. CulturalBench~\cite{chiu2025culturalbenchrobustdiversechallenging} tests broad everyday cultural knowledge across 45 countries through QA, and BLEnD~\cite{myung2025blendbenchmarkllmseveryday} evaluates everyday cultural knowledge in 13 languages across 16 regions. However, these benchmarks focus on general cultural literacy rather than deep domain expertise requiring professional-level judgment. SommBench complements this landscape by targeting a domain where expert knowledge is deeply shaped by cultural practices and linguistic discourse, spanning from objective factual knowledge (WTQA) through structured attribute prediction (WFC) to subjective expert judgment (FWP). Unlike broad cultural benchmarks, SommBench uses parallel multilingual content to directly measure cross-lingual consistency, and includes structured prediction and reasoning tasks beyond factual QA.

\begin{figure*}[!ht]
    \centering
    \includegraphics[width=\textwidth]{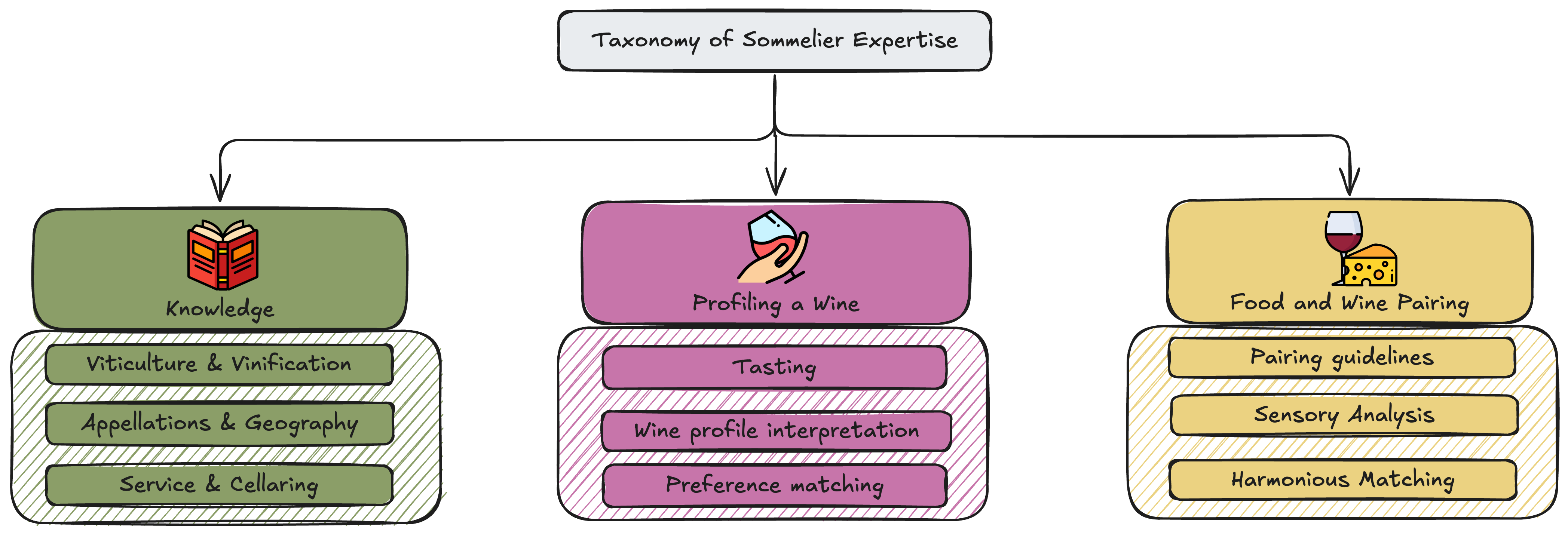}
    \caption{Taxonomy of Sommelier expertise, categorized into knowledge, profiling, and food wine pairing.}
    \label{fig:figure_2}
\end{figure*}

This approach is informed by recent work on defining and measuring culture in LLMs. The survey by \citet{adilazuarda-etal-2024-towards} notes that most research measures culture through proxies like region or values, and finds that domains like ``food and drink'' are largely unexplored. Our work also provides a practical application of theoretical frameworks like that of \citet{hershcovich-etal-2022-challenges}, who identify dimensions of cultural awareness in NLP, including shared "common ground" (facts), culturally relevant topics or "aboutness," and objectives or "values." SommBench is designed to probe these dimensions, testing whether factual common ground remains stable across languages and how models handle topics and values (e.g., food pairings) rooted in cultural practices.

Our work complements other expert benchmarks. While broad evaluations like MMMU~\cite{yue2023mmmu} test for general academic knowledge, and domain specific benchmarks like DrBenchmark~\cite{labrak-etal-2024-drbenchmark} or procedural ones like The Recipe Execution benchmark~\cite{nevens-etal-2024-benchmark} assess specialized, often monolingual, capabilities, SommBench offers a targeted benchmark for the cross-lingual coherence of culturally situated expertise. Within the wine domain specifically, prior work like WineSensed~\cite{bender2023learning} has focused on learning sensory representations from multimodal data. In contrast, our goal is to evaluate whether LLMs have already assimilated the complex, language inflected knowledge of a sommelier and can apply it consistently across different languages relying solely on textual representations of sensory properties such as taste, smell, and appearance.  

\section{SommBench}

SommBench is a multi-task and multilingual benchmark designed to evaluate language models on expert-level wine knowledge and their capabilities to emulate the behavior of a professional sommelier. It comprises three complementary tasks that capture different facets of expertise: \textit{Wine Theory Question-Answering} (WTQA, detailed in \S\ref{wtqa}) tests factual knowledge,  and \textit{Wine Features Completion} (WFC, \S\ref{wfc}) assesses the ability to reproduce accurate features of wine and \textit{Food \& Wine Pairing} (FWP, \S\ref{fwp}) evaluates the ability to judge food-wine pairings covering the taxonomy of sommelier \cite{manske2005understanding, parra2024towards} expertise displayed in Figure \ref{fig:figure_2}.

Crucially, SommBench provides an entirely new dataset, curated by an internationally recognized professional sommelier to ensure all tasks reflect authentic real-world challenges. While the underlying domain knowledge draws on publicly available sources (e.g.\ WSET~\footnote{\url{https://www.wsetglobal.com/}} curricula, wine retailer), the specific question-answer pairs with expert-crafted distractors, structured wine profiles, and sommelier pairing judgments are original artifacts that have not been publicly released prior to this work. This design is intentional: because the foundational knowledge may appear in pre-training corpora, SommBench tests whether models can apply and synthesise domain expertise in structured, expert level tasks rather than simply retrieve memorised facts providing a robust measure of generalization capabilities. 

\subsection{Wine Theory Question-Answering}\label{wtqa}

\paragraph{Data Collection}
The question-answer (QA) pairs were manually constructed by a master sommelier in English. The dataset comprises 128 question-answer pairs per language across 8 languages (English, Slovak, Swedish, Finnish, German, Danish, Italian, and Spanish), totalling 1,024 questions.

\paragraph{Data Curation}
Domain relevance and difficulty were validated by a master sommelier who assessed the entire question set for authenticity and expert-level complexity. Each question was authored as a complete four-option multiple-choice item by the master sommelier, who created both the correct answer and three distractor options simultaneously. The distractors were intentionally crafted to be plausible alternatives often reflecting common misconceptions.

\paragraph{Multilinguality}
The English QA set was initially translated using GPT-4.1 and then provided to native speakers for manual validation and correction into 7 target languages: \textit{Slovak, Swedish, Finnish, German, Danish, Italian, and Spanish}. Each language was validated by one dedicated native speaker fluent in the target language. The validators also conducted a verification pass with a focus on ensuring that all domain-specific nuances were accurately preserved across languages and their cultural interpretations.

\paragraph{The Task}
Wine Theory Question-Answering (WTQA) evaluates language models on their breadth and depth of wine knowledge, spanning fundamental concepts to expert-level oenological theory. This multilingual task tests whether models can accurately recall factual information that forms the foundation of sommelier expertise, knowledge traditionally acquired through formal certification programs like WSET (Wine \& Spirits Education Trust). The multilingual nature of this task adds a layer of complexity, as models must navigate domain-specific terminology that may not have direct translations across languages. For example, terms like "terroir" or "vendange tardive" carry specific meanings in wine contexts that require cultural and linguistic understanding beyond simple translation. We evaluate across multiple languages to assess models' ability to perform sommelier tasks in different linguistic contexts, assuming that wine expertise is often expected to be exercised in one's native language, where specific regional terminology is most naturally expressed. Example questions spanning all four difficulty levels are provided in Appendix~\ref{appendix:wtqa_examples}.

\paragraph{Evaluation}

The evaluation is handled using a generative approach where the model is presented with the question and all four multiple-choice answer options (A, B, C, D), and it is then prompted to generate its response. We instruct the model that its response should be a single letter (A, B, C, or D) and nothing else.
The generated output is subsequently parsed to extract the selected response letter ("A", "B", "C", or "D"). For reasoning models, the reasoning trace is ignored when extracting the response label.
The extracted label is then compared to the ground-truth response label. If the generated letter matches the correct one (e.g., the model generates "A" and the answer is "A"), it is marked as correct. Results are reported as accuracy scores for each language independently, together with an 'overall' column showing the average multilingual accuracy. The prompt template is provided in Appendix \ref{appendix:prompts}, Listing \cref{prompt:wtqa_en,prompt:wtqa_sk,prompt:wtqa_sv,prompt:wtqa_de,prompt:wtqa_dk,prompt:wtqa_it,prompt:wtqa_es,prompt:wtqa_fi}.

\subsection{Wine Features Completion}\label{wfc}
\paragraph{Data Collection}
A structured dataset of 1,000 wines was created by programmatically extracting information from major retailers and distributors' websites with a wide range of wine types. We parsed and normalised nine key attributes: type, sugar ($g/L$), alcohol (\%), country, region, grape varietals, dryness, body, and acidity.  These attributes represent the core characteristics essential for conducting professional analysis, much like a sommelier would. While retailer sourced data may carry positive framing in descriptive text, WFC targets structured factual attributes that are less susceptible to such bias than subjective tasting notes or marketing descriptions.

\paragraph{Data Curation}
Wine entries missing attributes were discarded during a comprehensive verification pass, ensuring every data point is fully populated and reliable. 

\paragraph{Multilinguality}
Wine Features Completion (WFC) is multilingual, covering the same set of languages as WTQA, supporting various languages for both prompt inputs and the corresponding output features. To ensure linguistic accuracy, English prompts and output features were translated using GPT-4.1 into all target languages and subsequently provided to one native speaker per language for manual validation and correction.

\paragraph{The Task}
The wine features a completion task designed to evaluate a model's ability to perform multilingual structured data generation. WFC challenges models to infer missing attributes within a wine's profile by synthesizing contextual clues from partially complete data, a capability crucial for real world applications like knowledge base completion and recommender systems with partial information. To systematically probe different completion capabilities, we employ a tiered masking strategy that creates evaluation subsets of increasing difficulty: First, a single mask is applied in $\sim$40\% of the cases, which tests the factual recall ability of the language model. Second, double-mask ($\sim$30\%): masks logical pairs (e.g., [region, country], [type, grapes], [type, region], [sugar, alcohol]) to assess simple inference. And lastly, triple-mask ($\sim$30\%) masks combinations of three attributes (e.g., [country, region, type], [alcohol, sugar, grapes], [dryness, acidity, body]) to evaluate complex compositional completion under high uncertainty. 

A key feature of WFC is its task structure, which explicitly tests multilingual generation from a language-independent input. While the input data is represented in a unified, canonical format, models are prompted to generate the completed wine profile in one of eight target languages (e.g., English, Slovak, German, Danish, Finnish, Swedish, Italian and Spanish). This output must be returned in a structured JSON format (Pydantic class style), testing the model's ability to map a canonical entity to its locale specific counterpart.

\paragraph{Evaluation}
We evaluate performance using distinct metrics tailored to attribute type: exact match for categorical fields (type, country, region, dryness, acidity and body) and mean absolute percentage error \cite{de2016mean} (MAPE) for numerical fields (alcohol, sugar). For these numerical attributes, a generated value is considered correct if its MAPE is $5$\% or less, and incorrect otherwise. Further details on the prompt template are provided in Appendix, Figure~\ref{prompt:wfc_judge}.

\subsection{Food \& Wine Pairing}\label{fwp}

\paragraph{Data Collection}
The dataset was constructed in collaboration with a professional sommelier. The sommelier created the positive instances by pairing wines with a diverse set of recipes. For the negative instances, an expert validated sampling strategy was employed: candidate wine-recipe pairs were first generated at random and then reviewed by the sommelier. Only pairs that were explicitly identified as 'negative' (i.e. bad pairings) were included in the dataset. This ensures that the negative class consists of genuinely unsuitable pairings, rather than random or neutral ones.

\paragraph{Data Curation}
To ensure consistency and account for the subjectivity inherent in wine pairing, one sommelier was responsible for labeling all instances (both positive and negative), thereby ensuring internal consistency and avoiding contradictory labels.

\paragraph{Monolinguality}
All Food \& Wine Pairing (FWP) data, including recipes and wine descriptions, is in English only. Unlike WTQA and WFC, where inputs are short structured fields amenable to translation, FWP requires full recipe descriptions with detailed ingredient lists and preparation steps. Translating these while preserving culinary nuance across eight languages would require substantial expert effort, and validating pairing judgments in each language would multiply the annotation burden.

\paragraph{The Task}
The food \& wine pairing task evaluates models on complex decision-making that requires domain expertise in both oenology, wine making and gastronomy. While food \& wine pairing recommendations are available online, these often reflect popular conventions rather than expert reasoning about flavor profiles, tannin structures, and sensory balance. This task tests whether models can internalize and apply the nuanced principles governing sensory interactions between food and wine, expertise typically acquired through years of specialized training. The FWP dataset comprises 1000 expert validated pairings in English only. Models assess whether a given wine-recipe combination constitutes a good pairing, responding with a binary ``Yes'' or ``No'' decision. This design directly measures the model's ability to distinguish harmonious from discordant pairings without allowing hedge responses. An example of a good pairing  (\checkmark) would be A5 Japanese Wagyu with sweet potato, orange, and collard greens served with Mouro Red 2018 from Alentejo in Portugal. The wine's smooth texture and succulent ripe flavors complement the meat's richness and the sweet notes found on the plate. On the other hand, a poor pairing example could be demonstrated with a spaghetti pasta recipe with creamy tomato sauce and chicken and served with an oaked Barossa Shiraz such as John Duval's ``Eligo'' Barossa Shiraz 2018. Here you have a sauce with acidity, sweetness and potentially a touch of umami and a delicate white meat. The intensity of the wine would take over the dish, while the intense oak spicing would clash aromatically with the delicate chicken and tomato flavours. 

\paragraph{Evaluation}
This task is evaluated as a binary classification, where models predict whether a given food-wine pairing is suitable (yes/no answers). Since both positive (yes) and negative (no) responses represent valid ground truth labels, we use the Matthews Correlation Coefficient (MCC) \cite{chicco2020advantages} as our primary evaluation metric. MCC ranges from -1 to +1 and provides a balanced assessment of binary classification quality by accounting for all confusion matrix categories (true positives, true negatives, false positives, and false negatives). Unlike the F1 score, which priorities the positive class and can be misleading when both classes are equally important, MCC treats both prediction outcomes symmetrically. This makes MCC particularly appropriate for our task, where correctly identifying bad pairings is as valuable as identifying good ones. We further report class-specific accuracy to identify potential biases toward positive or negative predictions. 

\subsection{Final Benchmark and Scoring}
\label{sec:sommbench_score}

The final benchmark comprises the three tasks described above, along with their corresponding, newly collected, datasets: 
\begin{itemize}
    \item WTQA: 128 question-answer pairs, totalling 1,024 items in 8 languages
    \item WFC: 1,000 wine entries for wine feature completion in 8 languages
    \item FWP: 1,000 monolingual food-wine pairings
\end{itemize}

To provide a single, comprehensive number for evaluating a large language model's capabilities as a sommelier, we introduce the SommBench Score. This score aggregates a model's performance across all tasks (WTQA, WFC, FWP) into a unified score. The aggregation method is designed to give equal weight to the different facets of expertise evaluated in the benchmark: multilingual factual knowledge, structured data completion, and culturally grounded judgment. The final score is calculated as the standard mean of the performance on each of the three tasks. As the FWP task is monolingual (English-only) by design, so English score is used. For the multilingual WTQA and WFC tasks, we use the mean score across all eight languages to directly measure cross-linguistic consistency. The final $Score$ is the arithmetic mean of these three component scores $S$, providing a holistic evaluation of sommelier expertise.
\begin{gather*}
    \text{SommBench-Score} = \frac{S_\mathrm{FWP} + \overline{S_\mathrm{WTQA}} + \overline{S_\mathrm{WFC}}}{3} \\
    \text{where} \quad \overline{S_\mathrm{WTQA}} = \frac{1}{|L|} \sum_{l \in L} S_\mathrm{WTQA, l} \\
    \text{and} \quad \overline{S_\mathrm{WFC}} = \frac{1}{|L|} \sum_{l \in L} S_\mathrm{WFC, l}
\end{gather*}

\section{Experiments}

We evaluated a comprehensive suite of 18 large language models on SommBench, including both leading closed-weights models and prominent open-weights alternatives. Evaluated closed-weights models are gpt-5 \cite{OpenAI2025GPT5}, gpt-4.1 family \cite{OpenAI2025GPT4_1} (gpt-4.1, gpt-4.1-mini, gpt-4.1-nano),gpt-4o family \cite{openai2024gpt4ocard} (gpt-4o, gpt-4o-mini), gemini family \cite{comanici2025gemini25pushingfrontier} (gemini-2.5-pro, gemini-2.5-flash, gemini-2.5-flash-lite) and grok-4 family \cite{xAI2025Grok4} (grok-4, grok-4-fast). Evaluated open weight models: gpt-oss models \cite{openai2025gptoss120bgptoss20bmodel} (gpt-oss-120, gpt-oss-20), qwen3 models \cite{yang2025qwen3technicalreport} (qwen3:30b, qwen3:8b), qwen2.5:3b \cite{qwen2.5}, gemma3:27b \cite{gemmateam2025gemma3technicalreport}, llama3.1:8b \cite{grattafiori2024llama3herdmodels}. The overall performance is summarized in Table \ref{tab:sommbench_results}, using the SommBench Score a holistic metric \ref{sec:sommbench_score}. Our results highlight a clear performance gap between closed and open-weights models. The top-performing model is gemini-2.5-flash, which achieves an overall SommBench Score of 0.65. It outperforms other strong closed-weights models, including gpt-4.1 (0.59) and gpt-5 (0.57). Among the open-weights models, qwen3:30b achieves the highest score of 0.51. The subsequent sections provide a detailed breakdown of model performance on each task, further examining their specific strengths and weaknesses. All models were evaluated in a zero-shot setting with the temperature set to 0 to ensure deterministic outputs. Reasoning was disabled for Qwen models, while gpt-oss models were tested at three reasoning intensity levels (low, medium, high). Open-weights models were served on 8$\times$A100 GPUs. Each configuration was evaluated in a single run.

\begin{table}[!ht]
\begin{center}
\resizebox{\columnwidth}{!}{
\begin{tabular}{lcccc}
\toprule 
\textbf{Model} & \textbf{WTQA} & \textbf{WFC} & \textbf{FWP} & \textbf{Score} \\
\midrule 
\multicolumn{5}{c}{\textit{Closed-Weights Models}} \\

gemini-2.5-flash & 0.9 & \textbf{0.63} & \textbf{0.39} & \textbf{0.65} \\
gpt-4.1 & 0.9 & 0.62 & 0.25 & 0.59 \\
gpt-4o & 0.9 & \textbf{0.63} & 0.19 & 0.57 \\
gemini-2.5-pro & 0.96 & 0.62 & 0.12 & 0.57 \\
gpt-5 & \textbf{0.97} & 0.57 & 0.17 & 0.57 \\
gpt-4.1-mini & 0.8 & 0.61 & 0.2 & 0.54 \\
grok-4 & 0.96 & 0.61 & 0.01 & 0.53 \\
grok-4-fast & 0.93 & 0.59 & 0.05 & 0.52 \\
gpt-4o-mini & 0.8 & 0.62 & 0.12 & 0.51 \\
gemini-2.5-flash-lite & 0.83 & 0.57 & 0.06 & 0.49 \\
gpt-4.1-nano & 0.73 & 0.52 & -0.02 & 0.41 \\
\midrule 
\multicolumn{5}{c}{\textit{Open-Weights Models}} \\

qwen3:30b & 0.84 & 0.48 & 0.2 & 0.51 \\
gemma3:27b & 0.76 & 0.52 & 0.23 & 0.50 \\
gpt-oss-120b (r=low) & 0.80 & 0.41 & 0.20 & 0.47 \\
gpt-oss-120b (r=medium) & 0.84 & 0.40 & 0.18 & 0.47 \\
gpt-oss:20b (r=low) & 0.67 & 0.33 & 0.25 & 0.42 \\
gpt-oss:20b (r=medium) & 0.72 & 0.38 & 0.16 & 0.42 \\
gpt-oss-120b (r=high) & 0.85 & 0.07 & 0.11 & 0.34 \\
qwen3:8b & 0.64 & 0.43 & -0.08 & 0.33 \\
llama3.1:8b & 0.53 & 0.44 & -0.01 & 0.32 \\
qwen2.5:3b & 0.48 & 0.24 & 0.1 & 0.27 \\
\bottomrule 
\end{tabular}
}
\end{center}
\caption{Performance of language models on SommBench. Models are evaluated using the SommBench score calculated from WFC, WTQA and FWP tasks. The Qwen models were employed without reasoning.}
\label{tab:sommbench_results}
\end{table}

\subsection{WTQA Results}

The wine theory question answering task evaluates a model's ability to recall expert level factual sommelier knowledge across eight different languages. The results, presented in Table~\ref{tab:wtqa_results}, indicate that the leading closed-source models possess a comprehensive and robust understanding of oenological theory. Top-performing models like grok-4 and gpt-5 achieve near-perfect accuracy, with scores often exceeding 95\% across the majority of tested languages. This demonstrates that state of the art LLMs have successfully assimilated a deep base of wine knowledge. A key finding from this task is the variance in cross-lingual consistency. While the top models maintain their high accuracy irrespective of the query language, many smaller and open-source models exhibit a performance degradation in non-English languages. This disparity is particularly stark for models like llama3.1:8b, which achieves an accuracy of 0.70 in English but plummets to 0.27 in Slovak and 0.44 in Swedish. A similar trend is observed for qwen3:8b, whose performance drops from 0.75 in English to 0.49 in Finnish or 0.59 in Danish. 

\begin{table}[!ht]
\begin{center}
\resizebox{\columnwidth}{!}{
\begin{tabular}{lcccccccc}
\toprule
\textbf{Model} & \textbf{EN} & \textbf{SK} & \textbf{SV} & \textbf{FI} & \textbf{DE} & \textbf{DA} & \textbf{IT} & \textbf{ES}\\
\midrule
\multicolumn{9}{c}{\textit{Closed-Weights Models}} \\
grok-4 & \textbf{0.98} & 0.96 & 0.97 & 0.95 & \textbf{0.97} & 0.97 & \textbf{0.98} & 0.94 \\
gpt-5 & \textbf{0.98} & \textbf{0.98} & \textbf{0.99} & 0.95 & \textbf{0.97} & \textbf{0.98} & \textbf{0.98} & 0.93 \\
gemini-2.5-pro & \textbf{0.98} & 0.95 & 0.98 & \textbf{0.96} & \textbf{0.97} & 0.96 & 0.96 & \textbf{0.95} \\
grok-4-fast & 0.94 & 0.94 & 0.93 & 0.91 & 0.94 & 0.92 & 0.94 & 0.91 \\
gemini-2.5-flash & 0.91 & 0.92 & 0.93 & 0.92 & 0.93 & 0.94 & 0.93 & 0.89 \\
gpt-4.1 & 0.91 & 0.88 & 0.92 & 0.91 & 0.91 & 0.93 & 0.91 & 0.85 \\
gpt-4o & 0.91 & 0.89 & 0.94 & 0.9 & 0.91 & 0.92 & 0.9 & 0.87 \\
gemini-2.5-flash-lite & 0.85 & 0.85 & 0.84 & 0.78 & 0.83 & 0.82 & 0.85 & 0.81 \\
gpt-4.1-mini & 0.9 & 0.73 & 0.84 & 0.8 & 0.85 & 0.81 & 0.77 & 0.74 \\
gpt-4o-mini & 0.84 & 0.77 & 0.79 & 0.7 & 0.83 & 0.81 & 0.81 & 0.81 \\
gpt-4.1-nano & 0.8 & 0.69 & 0.74 & 0.73 & 0.72 & 0.72 & 0.74 & 0.73\\
\midrule
\multicolumn{9}{c}{\textit{Open-Weights Models}} \\
gpt-oss-120b (r=low) & 0.81 & 0.73 & 0.81 & 0.81 & 0.83 & 0.79 & 0.81 & 0.80 \\
gpt-oss-120b (r=medium) & 0.86 & 0.82 & 0.88 & 0.81 & 0.84 & 0.84 & 0.84 & 0.85 \\
gpt-oss-120b (r=high) & 0.88 & 0.84 & 0.87 & 0.81 & 0.86 & 0.83 & 0.89 & 0.83 \\
qwen3:30b & 0.84 & 0.81 & 0.84 & 0.81 & 0.84 & 0.89 & 0.86 & 0.86 \\
gemma3:27b & 0.78 & 0.73 & 0.77 & 0.73 & 0.76 & 0.76 & 0.77 & 0.75 \\
gpt-oss:20b (r=low) & 0.75 & 0.59 & 0.68 & 0.62 & 0.68 & 0.63 & 0.71 & 0.67\\
gpt-oss:20b (r=medium) & 0.72 & 0.70 & 0.78 & 0.68 & 0.70 & 0.70 & 0.76 & 0.73\\
qwen3:8b & 0.75 & 0.54 & 0.66 & 0.49 & 0.68 & 0.59 & 0.73 & 0.69 \\
llama3.1:8b & 0.7 & 0.27 & 0.44 & 0.51 & 0.61 & 0.61 & 0.57 & 0.52 \\
qwen2.5:3b & 0.6 & 0.43 & 0.41 & 0.44 & 0.48 & 0.4 & 0.53 & 0.55 \\
\bottomrule
\end{tabular}%
}
\caption{Accuracy scores for language models on the Wine Theory Question-Answering (WTQA) task across eight languages (EN: English, SK: Slovak, SV: Swedish, FI: Finnish, DE: German, DA: Danish, IT: Italian, ES: Spanish). The Qwen models were employed without reasoning.}
\label{tab:wtqa_results}
\end{center}
\end{table}

\subsection{WFC Results}

The wine features a completion task that assesses the models' ability to infer missing wine attributes from a partial profile, testing multilingual structured generation. The results, shown in Table \ref{tab:wfc_results}, reveal a difference in cross-lingual consistency between model types. Top tier closed-source models such as gemini-2.5-flash, gemini-2.5-pro, and gpt-5 show stability across all eight languages, maintaining high performance with minimal variance. This suggests their internal knowledge of wine characteristics is largely language independent. In contrast, open-source models show a performance degradation when prompted in languages other than English. For instance, qwen3:30b's performance drops from 0.57 in English to 0.37 in Slovak, and llama3.1:8b declines from 0.53 in English to 0.36 in Finnish. A full breakdown of the accuracy of each attribute is provided in the Appendix in Table~\ref{tab:wfc_attribute_breakdown}.

\begin{table}[!ht]
\begin{center}
\resizebox{\columnwidth}{!}{
\begin{tabular}{lcccccccc} 
\toprule
\textbf{Model} & \textbf{EN} & \textbf{SK} & \textbf{SV} & \textbf{FI} & \textbf{DE} & \textbf{DA} & \textbf{IT} & \textbf{ES} \\ 
\midrule
\multicolumn{9}{c}{\textit{Closed-Weights Models}} \\ 
gpt-5 & 0.57 & 0.56 & 0.59 & 0.55 & 0.56 & 0.58 & 0.59 & 0.57 \\ 
gemini-2.5-pro & 0.62 & 0.61 & 0.62 & 0.62 & 0.62 & 0.63 & 0.62 & 0.61 \\ 
grok-4 & 0.62 & 0.6 & 0.59 & 0.62 & 0.59 & 0.61 & 0.6 & 0.62 \\ 
gemini-2.5-flash & \textbf{0.64} & \textbf{0.65} & 0.61 & \textbf{0.63} & 0.63 & 0.63 & 0.65 & 0.62 \\ 
gpt-4.1 & 0.61 & 0.64 & 0.62 & \textbf{0.63} & 0.62 & 0.63 & 0.62 & \textbf{0.63} \\ 
grok-4-fast & 0.59 & 0.6 & 0.58 & 0.57 & 0.61 & 0.62 & 0.57 & 0.58 \\ 
gem.-2.5-flash-lite & 0.56 & 0.61 & 0.55 & 0.56 & 0.59 & 0.56 & 0.55 & 0.6 \\ 
gpt-4o & 0.62 & 0.64 & \textbf{0.64} & 0.62 & \textbf{0.65} & \textbf{0.64} & \textbf{0.64} & 0.62 \\ 
gpt-4o-mini & 0.65 & 0.61 & 0.61 & \textbf{0.63} & 0.63 & 0.61 & 0.62 & 0.62 \\ 
gpt-4.1-mini & 0.6 & 0.62 & 0.61 & 0.61 & 0.62 & 0.62 & 0.59 & 0.6 \\ 
gpt-4.1-nano & 0.51 & 0.48 & 0.51 & 0.52 & 0.54 & 0.49 & 0.56 & 0.55 \\ 
\midrule 
\multicolumn{9}{c}{\textit{Open-Weights Models}} \\ 
gpt-oss-120b (r=low)  & 0.48 & 0.40 & 0.40 & 0.36 & 0.42 & 0.39 & 0.43 & 0.43 \\ 
gpt-oss-120b  (r=medium)& 0.47 & 0.39 & 0.39 & 0.36 & 0.42 & 0.35 & 0.44 & 0.42 \\ 
gpt-oss-120b  (r=high)& 0.02 & 0.09 & 0.05 & 0.03 & 0.07 & 0.09 & 0.10 & 0.11 \\ 
qwen3:30b & 0.57 & 0.37 & 0.45 & 0.44 & 0.51 & 0.47 & 0.51 & 0.49 \\ 
gemma3:27b & 0.6 & 0.45 & 0.53 & 0.48 & 0.51 & 0.51 & 0.53 & 0.51 \\ 
gpt-oss:20b (r=low)  & 0.43 & 0.26 & 0.35 & 0.24 & 0.36 & 0.32 & 0.33 & 0.38 \\
gpt-oss:20b (r=medium)  & 0.48 & 0.26 & 0.38 & 0.35 & 0.42 & 0.34 & 0.43 & 0.39 \\
qwen3:8b & 0.57 & 0.33 & 0.41 & 0.38 & 0.47 & 0.42 & 0.47 & 0.43 \\ 
llama3.1:8b & 0.53 & 0.34 & 0.43 & 0.36 & 0.46 & 0.46 & 0.45 & 0.45 \\ 
qwen2.5:3b & 0.25 & 0.18 & 0.27 & 0.22 & 0.24 & 0.24 & 0.26 & 0.27 \\ 
\bottomrule
\end{tabular}%
}
\end{center}
\caption{Wine feature completion results of language models across eight languages. Thinking is disabled for qwen models.}
\label{tab:wfc_results}
\end{table}

\subsection{FWP results}
\label{sec:fwp_experiments}

The food \& wine pairing task, designed to test nuanced, expert-level judgment, revealed substantial differences in model capabilities, as detailed in Table \ref{tab:fwp_results}. Model performance, measured by the Matthews Correlation Coefficient (MCC), ranged from -0.08 to 0.39. The top-performing model, gemini-2.5-flash, achieved an MCC of 0.39, demonstrating a moderate but reliable predictive ability. Several models scored below zero (e.g., llama3.1:8b, gpt-4.1-nano, and qwen3:8b), indicating that their predictions are less reliable than random chance. 

Results from Table  \ref{tab:fwp_results} reveal a positivity bias among many models. They show a strong tendency to approve pairings, regardless of their actual compatibility. This is particularly evident in models like gpt-4o-mini, which correctly identifies 90\% of good pairings (true positive rate) but only 18\% of bad pairings (true negative rate). This bias leads to a high number of incorrect recommendations. Conversely, a few models like qwen3:30b display a more conservative bias, being less likely to approve a pairing. Models with more balanced performance across both classes achieve higher scores. For example, gemini-2.5-flash balances its ability to identify good pairings (0.59 TPR) with its ability to spot bad ones (0.79 TNR). Similarly, the open-weights gemma3:27b achieves comparable performance on both classes (0.65 TPR and 0.58 TNR). 

\begin{table}[!ht]
\centering
\begin{tabular}{lccc}
\toprule
\textbf{Model} & \textbf{TPR} & \textbf{TNR} & \textbf{MCC} \\
\midrule
\multicolumn{4}{c}{\textit{Closed-Weights Models}} \\
gemini-2.5-flash & 0.59 & \textbf{0.79} & \textbf{0.39} \\
gpt-4.1 & 0.72 & 0.53 & 0.25 \\
gpt-4.1-mini & 0.78 & 0.40 & 0.20 \\
gpt-4o & 0.72 & 0.46 & 0.19 \\
gpt-5 & 0.58 & 0.59 & 0.17 \\
gpt-4o-mini & \textbf{0.90} & 0.18 & 0.12 \\
gemini-2.5-pro & 0.68 & 0.44 & 0.12 \\
gemini-2.5-flash-lite & 0.83 & 0.21 & 0.06 \\
grok-4-fast & 0.87 & 0.16 & 0.05 \\
grok-4 & 0.81 & 0.20 & 0.01 \\
gpt-4.1-nano & 0.81 & 0.17 & -0.02 \\
\midrule
\multicolumn{4}{c}{\textit{Open-Weights Models}} \\
gpt-oss:20b (r=low) & 0.74 & 0.50 & 0.25 \\
gemma3:27b & 0.65 & 0.58 & 0.23 \\
qwen3:30b & 0.49 & 0.70 & 0.20 \\
gpt-oss-120b (r=low) & 0.71 & 0.48 & 0.20 \\
gpt-oss-120b (r=medium) & 0.70 & 0.48 & 0.18 \\
gpt-oss:20b (r=medium) & 0.83 & 0.31 & 0.16 \\
gpt-oss-120b (r=high) & 0.77 & 0.33 & 0.11 \\
gwen2.5:3b & 0.04 & 0.99 & 0.10 \\
llama3.1:8b & 0.33 & 0.67 & -0.01 \\
qwen3:8b & 0.74 & 0.20 & -0.08 \\
\bottomrule
\end{tabular}
\caption{Performance of language models on the FWP task. Models are evaluated using True Positive Rate (TPR) to measure the identification of good pairings, True Negative Rate (TNR) for spotting bad pairings, and the Matthews Correlation Coefficient (MCC) for a balanced assessment. Thinking is disabled for qwen models.}
\label{tab:fwp_results}
\end{table}

\section{Discussion}

\paragraph{SommBench is challenging}
\label{sec:complexity}

Our results show that SommBench presents a challenging benchmark dataset for current large language models, with the top performing model, gemini-2.5-flash, achieving an overall score of only 0.65. To contextualize these scores, we consider uninformed baselines: random guessing on the four-way multiple-choice WTQA task yields 25\% accuracy, while a random binary classifier on the balanced FWP task yields an MCC of 0.0. This indicates that the benchmark is far from being solved and successfully evaluates a range of capabilities beyond simple fact recall. The complexity varies substantially across the three tasks. The wine theory question answering task appears to be the most manageable for current frontier models: gpt-5 (0.97), gemini-2.5-pro (0.96), and grok-4 (0.96) achieved near-perfect accuracy, which demonstrates a robust assimilation of specialized, factual oenological theory. The wine features completion task presents a moderate challenge. This task requires factual knowledge in conjunction with multilingual structured generation. Frontier models like gemini-2.5-flash and gpt-4o scored 0.63, suggesting that inferring missing attributes from partial, language independent data is a non-trivial capability. The food \& wine pairing task is clearly the most complex and serves as the primary differentiator between models. Performance in this task, which compares LLM judgment with sommelier judgment, was overall low. The highest score was a moderate 0.39 MCC from gemini-2.5-flash, while several models scored around the random baseline of 0.0 MCC (e.g., grok-4 at 0.01, llama3.1:8b at -0.01, and qwen3:8b at -0.08), indicating their predictions are no better than random chance.

\paragraph{Open versus closed-weights models}

\begin{figure}[hbtp]
    \centering
    \includegraphics[width=\columnwidth]{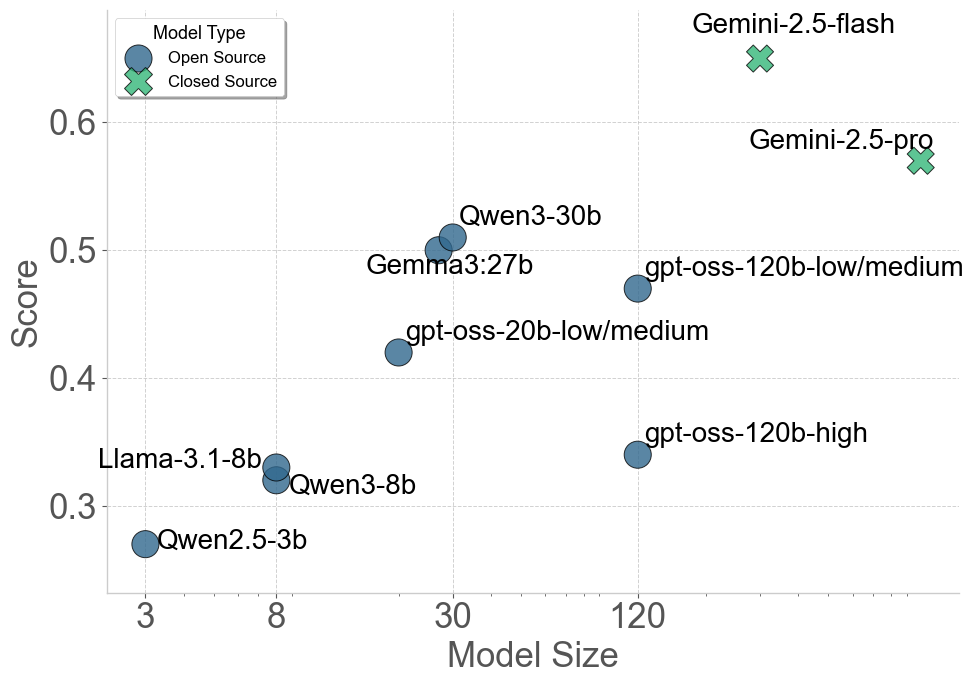}
    \caption{Scaling behaviour. SommBench score (y-axis) against number of parameters in billions (x-axis). gemini-2.5-flash and gemini-2.5-pro are plotted without a specific x-coordinate, as their parameter counts are undisclosed; they are included as performance reference points only.}
    \label{fig:figure_3}
\end{figure}

There is a clear difference in performance between closed-weights and open-weights models. All of the top-performing models in the benchmark are closed-weights models. The highest scoring open-weights model, qwen3:30b (0.51), was outperformed by nine of the eleven considered closed-weights models. This disparity is visually evident in Figure \ref{fig:figure_3}, which shows closed-source models clustering in the high-performance region while the performance of open-weights models scales with model size, following similar scaling laws as in \citet{kaplan2020scaling}.

\paragraph{Cross-lingual consistency} 

The gap between open and closed weights models is largely explained by a lack of cross-lingual consistency in open-weights models. While frontier models like gemini-2.5-pro and gpt-5 demonstrated stability across languages for WTQA and wine features completion, open-weights models showed a performance degradation in non-English languages. For example, in the WTQA task, llama3.1:8b's accuracy fell from 0.70 in English to 0.27 in Slovak. A similar decrease was observed for wine feature completion, where qwen3:30b scored 0.57 in English but only 0.37 in Slovak. This suggests that for many open-weights models, expert knowledge seems to be stored in a language-independent manner but tied to the linguistic context of the training data providing further evidence that language and cultural knowledge is allocated in different spaces of the models~\cite{namazifard2025isolating}.

\paragraph{Impact of reasoning} 
Analysis of reasoning levels using the gpt-oss models revealed a strong dependency on the specific task. Performance on the WTQA factual recall task increased steadily with reasoning intensity. Conversely, performance on the FWP judgment task peaked at the 'low' setting with higher levels being detrimental (reminiscent of ``overthinking''). The WFC task produced mixed results: the 'low' and 'medium' settings were comparable for the 120b model, but the 'medium' setting performed better for the 20b variant. Notably, the 'high' reasoning setting was catastrophic for both the FWP (0.11) and the WFC (0.07) tasks on the 120b model. The model engaged in flawed, complex reasoning for FWP and failed to produce the required structured output for WFC. This emphasizes that reasoning is highly contingent on specific task demands, such as factual retrieval versus constrained generation.

\paragraph{Should you trust an AI/LLM sommelier?} 
The answer depends on the specific task and model. For factual knowledge, top-tier closed-weights models like GPT-5 and gemini-2.5-pro demonstrate near perfect accuracy. However, for nuanced, more subjective tasks like food \& wine pairing, current models are not capable of emulating expert judgment. This task is the most complex, with the highest performing model, gemini-2.5-flash, achieving only a moderate (0.39 MCC) score. Several other models perform at the chance level. Many models exhibit a strong positivity bias (the tendency to approve rather than reject suggested pairings). Positivity bias is a well-documented phenomenon in LLMs, often discussed under the umbrella of sycophancy~\cite{perez-etal-2023-discovering,sharma2025understandingsycophancylanguagemodels}. However, unlike typical sycophancy studies where a model agrees with an explicitly stated user opinion, the SommBench FWP prompts are deliberately neutral: the model is simply asked ``Does the wine pair well with the recipe? Yes or No,'' with no user preference disclosed. This makes FWP a clean test of intrinsic approval bias rather than sycophantic behavior in general.

Two complementary hypotheses may explain this positivity bias. First, a training data hypothesis: wine-related content on the web skews positive — retail descriptions, food blogs, and pairing guides tend to recommend pairings rather than reject them. Models may therefore have learned a simplistic prior that food + wine = good pairing.
Second, an RLHF hypothesis: preference tuning rewards helpful and agreeable outputs, which could further bias models toward approval. Because our prompts contain no user opinion to agree with, classical sycophancy is unlikely to be the sole explanation; the bias likely also reflects the skewed distribution of pairing-related training data. The most extreme case is GPT-4o-mini, which predicted ``yes'' for 86\% of all pairings despite a perfectly balanced 50/50 ground truth, suggesting the model defaults to approval rather than exercising genuine pairing judgment. Of 1{,}500 negative (expert-rejected) pairings, GPT-4o-mini incorrectly approved 1{,}226, resembling an 82\% false-positive rate.
This suggests that frontier models could reliably automate factual wine descriptions, but deploying them as a virtual sommelier for meal recommendations is inadvisable. Current models too often fail to distinguish a genuinely good pairing from a poor one.

\section{Conclusion}
We introduced SommBench, an expert-curated benchmark in eight languages designed to evaluate the cultural and cross-lingual capabilities of large language models in the domain of sommelier expertise. It comprises three tasks: wine theory question answering, wine feature completion, and food-wine pairing.
 
Evaluating multiple LLMs on SommBench revealed key challenges in current state-of-the-art models. While leading models possess strong factual knowledge, many open-weight models show substantially lower performance in non-English languages, highlighting the need for more robust multilingual capabilities. Furthermore, we discovered a positivity bias in the food-wine pairing task, demonstrating that emulating subjective, expert-level judgment remains a major challenge --  even for frontier language models.
 
SommBench highlights three critical challenges for globally deployed LLMs: cross-lingual consistency in culturally-grounded knowledge, the gap between factual recall and expert judgment, and the multilingual capabilities required for culturally situated domains. Addressing these challenges is crucial for developing language models that can be trusted across linguistic and cultural contexts.

\paragraph{Future Work}
Building on our findings, we identify four directions for extending SommBench. (1) Collecting annotations from a diverse panel of certified sommeliers to measure inter-annotator agreement and derive a consensus-based ground truth. (2) Making the food-wine pairing task fully multilingual by providing complete recipe descriptions in all eight target languages. (3) Enriching wine feature completion with subjective sensory descriptors, such as tasting notes and aroma profiles. (4) Moving beyond binary classification to a generative pairing task in which models recommend wines and justify their choices in natural language.

\section{Limitations}
\begin{itemize}
    \item All tasks are based on the annotations of a single master sommelier, and no formal inter-annotator agreement study was conducted. This single-annotator design was a deliberate methodological choice: wine evaluation is inherently subjective, and using a single consistently calibrated expert eliminates the noise that arises from reconciling conflicting opinions across annotators, a well-known challenge in subjective annotation tasks. The benchmark therefore measures alignment with a coherent, expert-level standard rather than an aggregated consensus, which we argue is more appropriate for a domain where even certified professionals routinely disagree on pairings and tasting descriptors. We acknowledge this as a limitation and plan to extend future iterations with annotations from a diverse panel of sommeliers, enabling formal inter-annotator agreement analysis and a more robust, consensus-based ground truth.
    \item The FWP task is limited to English. Unlike the structured inputs of WTQA and WFC, FWP relies on full recipe descriptions whose translation and re-validation across eight languages would require considerable expert effort. This limits the benchmark's ability to evaluate cross-lingual and culturally-adaptive reasoning in the context of food-wine pairing, and extending FWP multilingually is a priority for future work.
    \item The WFC task focuses on objective, structured attributes (e.g., alcohol content, country of origin, acidity level) and does not evaluate a model's ability to generate the rich, subjective sensory descriptors, such as tasting notes and aroma profiles, that are fundamental to sommelier expertise.
    \item The benchmark currently covers eight European languages, selected to represent several major wine-producing and wine-consuming cultures. However, this scope is not globally representative. Expanding to additional languages, particularly from wine regions in the Americas, Africa, and Asia, would be a valuable step toward a truly global evaluation of wine expertise.
\end{itemize}

\section{Ethical Considerations}
We have developed a multilingual sommelier benchmark. The benchmark has been developed in conjunction with a recognized, award-winning sommelier and with native speaker validation in the respective languages. We do not expect that this benchmark would incur any new risks (other than, potentially, mispredicted wine \& food pairings), but rather contributes to evaluating how fairly language model capabilities are distributed across languages and cultures. SommBench is intended strictly as a research benchmark and should not be used to replace the expertise of trained sommeliers. Given the limitations observed in the food-wine pairing task, results should not be used to market AI systems as reliable wine advisors. As the dataset concerns the alcohol domain, responsible deployment considerations should be taken into account in any downstream application. The dataset does not contain any personally identifiable information.

\section{Author Contributions}
We list author contributions according to the Contributor Roles Taxonomy (CRediT)\footnote{\url{https://credit.niso.org}}:

\begin{itemize}
    \item \textbf{Conceptualization:} W.~Brach, L.~Galke Poech
    \item \textbf{Data curation:} J.~Dupouy (en), W.~Brach (en, sk), T.~Bedej (en), J.~Pichna (fi), L.~Galke Poech (de), E.~Saarensilta (fi), J.~Nielsen (dk), G.~Barmina (it), A.~Blasi N\'{u}\~{n}ez (es)
    \item \textbf{Formal analysis:} W.~Brach, L.~Galke Poech
    \item \textbf{Investigation:} W.~Brach, J.~Nielsen
    \item \textbf{Project administration:} W.~Brach
    \item \textbf{Software:} W.~Brach
    \item \textbf{Writing -- original draft:} W.~Brach, L.~Galke Poech.
    \item \textbf{Writing -- review \& editing:} W.~Brach, J.~Pichna, L.~Galke Poech, J.~Nielsen, G.~Barmina, A.~Blasi N\'{u}\~{n}ez, J.~Pichna, E.~Saarensilta
    \item \textbf{Funding acquisition:} K. Košťál, M.~Ries, P.~Schneider-Kamp
    \item \textbf{Supervision:} L.~Galke Poech, K. Košťál, M.~Ries, P.~Schneider-Kamp
\end{itemize}

\section{Acknowledgements}

This work was supported by the Science Grant Agency - project VEGA 1/0300/25.
This work was further supported in parts by the strategic initiative on Danish Foundation Models.

\section*{References}
\bibliographystyle{lrec2026-natbib}
\bibliography{lrec2026-example}

\clearpage
\newcommand{\appendixfigurecaption}[2]{%
  \refstepcounter{figure}\label{#1}%
  \begin{center}\small Figure~\thefigure: #2\end{center}%
}
\newcommand{\appendixtablecaption}[2]{%
  \refstepcounter{table}\label{#1}%
  \begin{center}\small Table~\thetable: #2\end{center}%
}
\appendix
\section{Prompts}
\label{appendix:prompts}

This section presents the prompt templates used for each task. WTQA prompts are shown in all eight languages, followed by the FWP and WFC prompts.

Across all tasks, prompts were designed to minimise variance unrelated to domain knowledge. We therefore kept the instruction format stable, used explicit output constraints, and avoided additional chain-of-thought style guidance that could advantage some model families over others. The aim was to measure sommelier competence rather than prompt-following creativity.

For WTQA, each language version is a close translation of the same underlying instruction template. Native-speaker validators checked not only grammatical correctness but also whether wine-specific terminology retained the intended register and meaning. This is especially important in a domain where terms such as appellations, production methods, and tasting vocabulary are often borrowed across languages or only partially translated in practice.

For WFC and FWP, the prompts were likewise kept intentionally direct. In WFC, the structured output requirement tests whether models can reliably map partial evidence into a canonical schema. In FWP, the binary decision format prevents hedging and forces a clear judgment, making it easier to study approval bias and disagreement with expert labels.

\noindent\fbox{%
\begin{minipage}{\dimexpr\columnwidth-2\fboxsep-2\fboxrule\relax}
\texttt{\small
Act as an expert sommelier. \\
Your task is to answer the following multiple-choice question. \\
Your response MUST be a single letter (A, B, C, or D) and nothing else. \\
\\
Question: \{question\} \\
\\ 
Options: \\
(A) \{a\} \\ 
(B) \{b\} \\
(C) \{c\} \\
(D) \{d\} \\
\\
Correct Answer (A, B, C, or D):
}
\end{minipage}
}
\appendixfigurecaption{prompt:wtqa_en}{Prompt used for English WTQA.}

\noindent\fbox{%
\begin{minipage}{\dimexpr\columnwidth-2\fboxsep-2\fboxrule\relax}
\texttt{\small
Si expert someliér. \\
Tvojou úlohou je odpovedať na nasledujúcu otázku s možnosťou výberu z viacerých odpovedí. \\
Tvoja odpoveď MUSÍ byť jediné písmeno (A, B, C alebo D) a nič iné. \\
\\
Otázka: \{question\} \\
\\ 
Možnosti: \\
(A) \{a\} \\
(B) \{b\} \\
(C) \{c\} \\
(D) \{d\} \\
\\
Správna odpoveď je (A, B, C alebo D):
}
\end{minipage}
}
\appendixfigurecaption{prompt:wtqa_sk}{Prompt used for Slovak WTQA.}

\noindent\fbox{%
\begin{minipage}{\dimexpr\columnwidth-2\fboxsep-2\fboxrule\relax}
\texttt{\small
Optræd som en ekspert-sommelier. \\
Din opgave er at besvare følgende multiple choice-spørgsmål. \\
Dit svar SKAL være et enkelt bogstav (A, B, C eller D) og intet andet. \\
\\
Spørgsmål: \{question\} \\
\\
Valgmuligheder: \\
(A) \{a\} \\
(B) \{b\} \\
(C) \{c\} \\
(D) \{d\} \\
\\
Korrekt svar (A, B, C eller D):
}
\end{minipage}
}
\appendixfigurecaption{prompt:wtqa_dk}{Prompt used for Danish WTQA.}

\noindent\fbox{%
\begin{minipage}{\dimexpr\columnwidth-2\fboxsep-2\fboxrule\relax}
\texttt{\small
Verhalten Sie sich wie ein erfahrener Sommelier. \\
Ihre Aufgabe ist es, die folgende Multiple-Choice-Frage zu beantworten. \\
Ihre Antwort MUSS aus einem einzigen Buchstaben (A, B, C oder D) bestehen und darf nichts anderes enthalten. \\
\\
Frage: \{question\} \\
\\
Optionen: \\
(A) \{a\} \\
(B) \{b\} \\
(C) \{c\} \\
(D) \{d\} \\
\\
Richtige Antwort (A, B, C oder D):
}
\end{minipage}
}
\appendixfigurecaption{prompt:wtqa_de}{Prompt used for German WTQA.}

\noindent\fbox{%
\begin{minipage}{\dimexpr\columnwidth-2\fboxsep-2\fboxrule\relax}
\texttt{\small
Agisci come un sommelier esperto. \\
Il tuo compito è rispondere alla seguente domanda a scelta multipla. \\
La tua risposta DEVE essere una sola lettera (A, B, C o D) e nient'altro. \\
\\
Domanda: \{question\} \\
\\
Opzioni: \\
(A) \{a\} \\
(B) \{b\} \\
(C) \{c\} \\
(D) \{d\} \\
\\
Risposta corretta (A, B, C o D):
}
\end{minipage}
}
\appendixfigurecaption{prompt:wtqa_it}{Prompt used for Italian WTQA.}

\noindent\fbox{%
\begin{minipage}{\dimexpr\columnwidth-2\fboxsep-2\fboxrule\relax}
\texttt{\small
Actúa como un sumiller experto. \\
Tu tarea consiste en responder a la siguiente pregunta de opción múltiple. \\
Su respuesta DEBE ser una sola letra (A, B, C o D) y nada más. \\
\\
Pregunta: \{question\} \\
\\
Opciones: \\
(A) \{a\} \\
(B) \{b\} \\
(C) \{c\} \\
(D) \{d\} \\
\\
Respuesta correcta (A, B, C o D):
}
\end{minipage}
}
\appendixfigurecaption{prompt:wtqa_es}{Prompt used for Spanish WTQA.}

\noindent\fbox{%
\begin{minipage}{\dimexpr\columnwidth-2\fboxsep-2\fboxrule\relax}
\texttt{\small
Agera som en expert-sommelier. \\
Din uppgift är att svara på följande flervalsfråga. \\
Ditt svar MÅSTE bestå av en enda bokstav (A, B, C eller D) och inget annat. \\
\\
Fråga: \{question\} \\
\\
Alternativ: \\
(A) \{a\} \\
(B) \{b\} \\
(C) \{c\} \\
(D) \{d\} \\
\\
Rätt svar (A, B, C eller D):
}
\end{minipage}
}
\appendixfigurecaption{prompt:wtqa_sv}{Prompt used for Swedish WTQA.}

\noindent\fbox{%
\begin{minipage}{\dimexpr\columnwidth-2\fboxsep-2\fboxrule\relax}
\texttt{\small
Toimi sommelier-asiantuntijana. \\
Tehtävänäsi on vastata seuraavaan monivalintakysymykseen. \\
Vastauksesi on oltava yksi kirjain (A, B, C tai D) eikä mitään muuta. \\
\\
Kysymys: \{question\} \\
\\
Vaihtoehdot: \\
(A) \{a\} \\
(B) \{b\} \\
(C) \{c\} \\
(D) \{d\} \\
\\
Oikea vastaus (A, B, C tai D):
}
\end{minipage}
}
\appendixfigurecaption{prompt:wtqa_fi}{Prompt used for Finnish WTQA.}

The following prompt was used for the Food \& Wine Pairing (FWP) binary classification task.

\noindent\fbox{%
\begin{minipage}{\dimexpr\columnwidth-2\fboxsep-2\fboxrule\relax}
\texttt{\small
Act as an expert sommelier. \\
Your task is to evaluate the pairing of a given wine and recipe. \\
Your response MUST be Yes or No and nothing else. \\~\\
Recipe: \{recipe\} \\
Wine: \{wine\} \\
Does the wine pair well with the recipe? Yes or No:
}
\end{minipage}
}
\appendixfigurecaption{prompt:fwp_judge}{Prompt used for the Food \& Wine Pairing task.}

The following prompt was used for the Wine Feature Classification (WFC) structured extraction task.

\noindent\fbox{%
\begin{minipage}{\dimexpr\columnwidth-2\fboxsep-2\fboxrule\relax}
\texttt{\small
Analyze the following wine description: \\
\{passage\} \\~\\
Based on this text, populate ALL fields of the required JSON structure. \\
For any attributes not explicitly mentioned, predict the most likely value based on the other information provided, ensuring the predicted value adheres to the required data type and enum constraints. \\~\\
Required JSON fields: \\
- type: "red" | "white" | "rose" | "sparkling" | "dessert" | "fortified" \\
- sugar: float (residual sugar in g/L) \\
- alcohol: float (alcohol content in \%) \\
- country: str (production country) \\
- region: str (geographical region or appellation) \\
- grapes: list[str] (primary grape varietals) \\
- dryness: "dry" | "medium dry" | "medium sweet" | "sweet" \\
- body: "light bodied" | "medium bodied" | "full bodied" \\
- acidity: "slightly acidic" | "medium acidic" | "acidity" | "very acidic"
}
\end{minipage}
}
\appendixfigurecaption{prompt:wfc_judge}{Prompt used for the Wine Feature Completion task. The model receives a partial wine description and must populate all fields of the required JSON structure.}

\section{Example WTQA Questions}
\label{appendix:wtqa_examples}

Table~\ref{tab:wtqa_examples} shows example questions from the Wine Theory Question-Answering task, with two questions per difficulty level.

These examples illustrate how difficulty increases from basic terminology to highly specialised oenological knowledge. The progression is important for interpreting the main results: very strong aggregate WTQA performance does not only reflect success on introductory concepts, but also competence on questions that require familiarity with region-specific terminology, vinification procedures, and certification-style theory.

\begin{enumerate}[leftmargin=*, label=\textbf{Level \arabic*:}, itemsep=0.5em]
    \item \textit{What does ``spumante'' mean?} Options: (A) Sparkling; (B) Still; (C) Semi-sparkling; (D) Flat. Correct answer: A. \\
    \textit{What does ``frizzante'' mean?} Options: (A) Still; (B) Sparkling; (C) Dry; (D) Lightly sparkling. Correct answer: D.
    \item \textit{What is the term for removing sediment from Champagne?} Options: (A) Disgorgement; (B) Riddling; (C) Dosage; (D) Assemblage. Correct answer: A. \\
    \textit{Where was Pinotage developed?} Options: (A) Australia; (B) South Africa; (C) New Zealand; (D) Chile. Correct answer: B.
    \item \textit{What is b\^{a}tonnage?} Options: (A) Stirring the lees; (B) Punching down the cap; (C) Racking the wine; (D) Fining the wine. Correct answer: A. \\
    \textit{What are the four main styles of Madeira (driest to sweetest)?} Options: (A) Tinta Negra, Verdelho, Bual, Malmsey; (B) Verdelho, Sercial, Malmsey, Bual; (C) Malmsey, Bual, Verdelho, Sercial; (D) Sercial, Verdelho, Bual, Malmsey. Correct answer: D.
    \item \textit{What is the German term for noble rot?} Options: (A) Trockenbeerenauslese; (B) Sp\"{a}tlese; (C) Edelf\"{a}ule; (D) Beerenauslese. Correct answer: C. \\
    \textit{What vine training system is used in Alsace?} Options: (A) Pergola; (B) Bush vine; (C) Cordon; (D) Guyot. Correct answer: D.
\end{enumerate}
\appendixtablecaption{tab:wtqa_examples}{Example questions from the WTQA task across all four difficulty levels. Level~1 covers basic wine terminology, Level~2 tests intermediate knowledge, Level~3 requires advanced understanding, and Level~4 targets expert-level oenological knowledge.}

\section{Dataset Statistics}
\label{appendix:dataset_stats}

Tables~\ref{tab:dataset_stats_wtqa} and~\ref{tab:dataset_stats_wfc} summarise the composition of each benchmark task.

We include these statistics to make the benchmark design more transparent and to contextualise the reported model scores. The WTQA distribution is intentionally skewed toward intermediate and advanced questions, reflecting the fact that professional wine knowledge extends well beyond basic terminology. Likewise, the WFC distribution follows the empirical availability of retailer data rather than an artificially balanced sampling scheme, making the task closer to practical catalogue-completion settings.

\begin{center}
\small
\begin{tabular}{|l|r|l|}
\hline
\textbf{Level} & \textbf{Count} & \textbf{Description} \\
\hline
Level 1 & 8 & Basic wine terminology \\
\hline
Level 2 & 33 & Intermediate knowledge \\
\hline
Level 3 & 66 & Advanced understanding \\
\hline
Level 4 & 21 & Expert-level oenology \\
\hline
\textbf{Total} & \textbf{128} & \textbf{per language (1,024 across 8)} \\
\hline
\end{tabular}
\end{center}
\appendixtablecaption{tab:dataset_stats_wtqa}{WTQA question distribution by difficulty level.}

\begin{center}
\small
\begin{tabular}{|l|r|}
\hline
\textbf{Wine Type} & \textbf{Count} \\
\hline
White & 283 \\
\hline
Sparkling & 270 \\
\hline
Red & 231 \\
\hline
Rosé & 161 \\
\hline
Dessert / Fortified & 55 \\
\hline
\textbf{Total} & \textbf{1,000} \\
\hline
\end{tabular}
\end{center}
\appendixtablecaption{tab:dataset_stats_wfc}{WFC wine profile distribution by type. The five most represented countries are France (346), Italy (182), Spain (138), Germany (76), and Austria (44).}

For the WFC masking strategy, approximately 40\% of profiles have a single attribute masked, 30\% have two attributes masked, and 30\% have three attributes masked, as described in the main text.

The FWP dataset is perfectly balanced with 500 positive (good pairing) and 500 negative (bad pairing) examples.

This balance is particularly important for interpreting FWP scores. Because the task contains equal numbers of positive and negative examples, models cannot obtain strong results simply by over-predicting one class. As a consequence, the gap between true positive and true negative rates offers a direct view into whether a model behaves like a cautious evaluator or defaults to approving pairings.

\subsection{Example WFC Entries}

Table~\ref{tab:wfc_examples} shows three representative wine profiles from the WFC dataset. Fields marked with \texttt{[?]} are masked and must be predicted by the model.

\begin{itemize}[leftmargin=*, itemsep=0.4em]
    \item \textbf{Barone Pizzini Animante Franciacorta Dosaggio Zero}: type = sparkling; sugar = 1.0 g/L; alcohol = 12.0\%; country = Italy; region = Lombardy; grapes = Pinot Blanc, Pinot Noir, Chardonnay; dryness = dry; acidity = \texttt{[?]}; body = full bodied.
    \item \textbf{Riunite Il Fojonco Lambrusco}: type = red; sugar = 41.0 g/L; alcohol = 8.0\%; country = Italy; region = Emilia-Romagna; grapes = Lambrusco Grasparossa; dryness = medium sweet; acidity = \texttt{[?]}; body = medium bodied.
    \item \textbf{St.\ Stephan's Crown Tokaji Aszú 5 Puttonyos 2019}: type = dessert; sugar = 154.0 g/L; alcohol = 12.0\%; country = Hungary; region = Tokaj-Hegyalja; grapes = Furmint; dryness = sweet; acidity = acidic; body = \texttt{[?]}.
\end{itemize}
\appendixtablecaption{tab:wfc_examples}{Example wine profiles from the WFC dataset. Models receive profiles with selected attributes replaced by \texttt{[?]} and must predict the missing values.}

\subsection{Example FWP Entries}

Table~\ref{tab:fwp_examples} shows representative food--wine pairings from the FWP dataset, including both expert-validated good pairings and bad pairings.

\begin{itemize}[leftmargin=*, itemsep=0.4em]
    \item \textbf{yes}: Riesling Blend, Domaine Weinbach, Alsace 2023 --- Steamed lobster; kumquat and charred cucumber, spiced shellfish-citrus broth.
    \item \textbf{yes}: Amontillado Sherry, Lustau, Andalucia --- Slowly baked patty pan filled with guajilla and ancho pepper; mole sauce.
    \item \textbf{no}: Ilramato Pinot Grigio 2022 --- Wild rabbit, langoustine and Jerusalem artichoke crumble with wild garlic crust.
    \item \textbf{no}: Arthur Metz Vin d'Alsace 2021 --- Mini Christmas cakes.
\end{itemize}
\appendixtablecaption{tab:fwp_examples}{Example entries from the FWP dataset. Good pairings (yes) are sourced from sommelier recommendations; bad pairings (no) are deliberately mismatched combinations validated by domain experts.}

\section{Full results}
\label{appendix:full_results}

This section provides extended results for the FWP and WFC tasks. We first report full FWP metrics including F1 scores along with a visual comparison, followed by a per-attribute breakdown of WFC accuracy.

The appendix tables make three recurring patterns easier to see. First, frontier models are already very strong on factual recall, but that strength does not transfer cleanly to judgment-heavy pairing decisions. Second, WFC performance is uneven across attributes: discrete fields such as wine type and country are generally easier than body, acidity, and sugar. Third, several models achieve acceptable overall scores while still displaying pronounced class imbalance or weak calibration, which is why the per-metric view remains informative even when the main paper already reports aggregate results.

\begin{center}
\resizebox{\columnwidth}{!}{%
\begin{tabular}{|l|c|c|c|c|}
\hline
\textbf{Model} & \textbf{TPR} & \textbf{TNR} & \textbf{F1} & \textbf{MCC} \\
\hline
gemini-2.5-flash & 0.59 & 0.79 & 0.69 & 0.39 \\
\hline
gpt-4.1 & 0.72 & 0.53 & 0.62 & 0.25 \\
\hline
gemma3:27b & 0.65 & 0.58 & 0.61 & 0.23 \\
\hline
qwen3:30b & 0.49 & 0.70 & 0.59 & 0.20 \\
\hline
gpt-4.1-mini & 0.78 & 0.40 & 0.58 & 0.20 \\
\hline
gpt-4o & 0.72 & 0.46 & 0.59 & 0.19 \\
\hline
gpt-oss:20b & 0.76 & 0.43 & 0.58 & 0.19 \\
\hline
gpt-5 & 0.58 & 0.59 & 0.58 & 0.17 \\
\hline
gpt-oss-120b & 0.69 & 0.46 & 0.57 & 0.15 \\
\hline
gpt-4o-mini & 0.90 & 0.18 & 0.47 & 0.12 \\
\hline
gemini-2.5-pro & 0.68 & 0.44 & 0.55 & 0.12 \\
\hline
gemini-2.5-flash-lite & 0.83 & 0.21 & 0.47 & 0.06 \\
\hline
grok-4-fast & 0.87 & 0.16 & 0.45 & 0.05 \\
\hline
grok-4 & 0.81 & 0.20 & 0.45 & 0.01 \\
\hline
llama3.1:8b & 0.33 & 0.67 & 0.48 & -0.01 \\
\hline
gpt-4.1-nano & 0.81 & 0.17 & 0.43 & -0.02 \\
\hline
qwen3:8b & 0.74 & 0.20 & 0.43 & -0.08 \\
\hline
\end{tabular}%
}
\end{center}
\appendixtablecaption{tab:full_results_fwp}{Performance of language models on the FWP task. Models are evaluated using True Positive Rate (TPR), True Negative Rate (TNR), F1 score, and Matthews Correlation Coefficient (MCC).}

\subsection{Per-Attribute WFC Breakdown}

Table~\ref{tab:wfc_attribute_breakdown} shows the per-attribute WFC breakdown, split into two narrow appendix tables for readability. Discrete attributes such as wine type and country of origin are predicted near-perfectly by most models, whereas continuous or subjective attributes sugar, body, and acidity remain the most challenging.

\begin{center}
\resizebox{\columnwidth}{!}{
\begin{tabular}{lccccc}
\toprule
\textbf{Model} & \textbf{Type} & \textbf{Country} & \textbf{Region} & \textbf{Grapes} & \textbf{Dry.} \\
\midrule
\multicolumn{6}{l}{\textit{Closed-Weights}} \\
gpt-5 & \textbf{1.00} & \textbf{1.00} & 0.43 & 0.81 & \textbf{0.63} \\
gemini-2.5-pro & 0.99 & 0.99 & 0.77 & 0.87 & 0.57 \\
grok-4 & \textbf{1.00} & \textbf{1.00} & 0.68 & 0.82 & \textbf{0.63} \\
gemini-2.5-flash & \textbf{1.00} & \textbf{1.00} & 0.85 & 0.86 & 0.59 \\
gpt-4.1 & \textbf{1.00} & \textbf{1.00} & 0.81 & 0.84 & 0.56 \\
grok-4-fast & 0.99 & \textbf{1.00} & 0.58 & 0.79 & \textbf{0.63} \\
gem.-2.5-flash-lite & \textbf{1.00} & 0.98 & 0.68 & 0.84 & 0.42 \\
gpt-4o & \textbf{1.00} & \textbf{1.00} & \textbf{0.87} & \textbf{0.92} & 0.54 \\
gpt-4o-mini & 0.98 & \textbf{1.00} & 0.68 & 0.87 & 0.55 \\
gpt-4.1-mini & 0.97 & \textbf{1.00} & 0.72 & 0.89 & 0.55 \\
gpt-4.1-nano & 0.83 & 0.98 & 0.54 & 0.83 & 0.35 \\
\midrule
\multicolumn{6}{l}{\textit{Open-Weights}} \\
gpt-oss-120b (r=low) & 0.32 & 0.13 & 0.54 & \textbf{0.86} & 0.64 \\
gpt-oss-120b (r=medium) & 0.37 & 0.15 & \textbf{0.55} & \textbf{0.86} & 0.60 \\
gpt-oss-120b (r=high) & 0.06 & 0.01 & 0.12 & 0.15 & 0.09 \\
qwen3:30b & 0.41 & \textbf{0.76} & \textbf{0.55} & 0.84 & 0.52 \\
gemma3:27b & \textbf{0.87} & 0.59 & 0.47 & 0.80 & \textbf{0.70} \\
gpt-oss:20b (r=low) & 0.26 & 0.14 & 0.51 & 0.69 & 0.50 \\
gpt-oss:20b (r=medium) & 0.32 & 0.09 & 0.49 & 0.77 & 0.57 \\
qwen3:8b & 0.48 & 0.67 & 0.47 & 0.77 & 0.51 \\
llama3.1:8b & 0.45 & 0.61 & 0.38 & 0.74 & 0.65 \\
qwen2.5:3b & 0.24 & 0.21 & 0.05 & 0.43 & 0.38 \\
\bottomrule
\end{tabular}%
}
\end{center}
\appendixtablecaption{tab:wfc_attribute_breakdown}{Per-attribute WFC accuracy for structural attributes (type, country, region, grapes, and dryness), aggregated across all eight languages. Best values per group are in bold.}

\begin{center}
\resizebox{\columnwidth}{!}{
\begin{tabular}{lcccc}
\toprule
\textbf{Model} & \textbf{Body} & \textbf{Acid.} & \textbf{Alc.} & \textbf{Sugar} \\
\midrule
\multicolumn{5}{l}{\textit{Closed-Weights}} \\
gpt-5 & 0.19 & 0.12 & 0.91 & 0.06 \\
gemini-2.5-pro & 0.24 & 0.13 & 0.91 & 0.10 \\
grok-4 & 0.18 & 0.06 & 0.88 & \textbf{0.21} \\
gemini-2.5-flash & 0.21 & \textbf{0.18} & 0.87 & 0.12 \\
gpt-4.1 & 0.15 & \textbf{0.18} & \textbf{0.94} & 0.14 \\
grok-4-fast & 0.20 & 0.12 & 0.82 & 0.17 \\
gem.-2.5-flash-lite & 0.23 & 0.17 & 0.72 & 0.12 \\
gpt-4o & 0.18 & \textbf{0.18} & 0.88 & 0.12 \\
gpt-4o-mini & \textbf{0.26} & 0.17 & 0.93 & 0.17 \\
gpt-4.1-mini & 0.23 & \textbf{0.18} & 0.79 & 0.15 \\
gpt-4.1-nano & 0.21 & 0.16 & 0.67 & 0.12 \\
\midrule
\multicolumn{5}{l}{\textit{Open-Weights}} \\
gpt-oss-120b (r=low) & 0.18 & 0.15 & \textbf{0.71} & 0.19 \\
gpt-oss-120b (r=medium) & 0.19 & 0.15 & 0.60 & 0.18 \\
gpt-oss-120b (r=high) & 0.03 & 0.05 & 0.05 & 0.05 \\
qwen3:30b & 0.19 & 0.16 & 0.69 & 0.18 \\
gemma3:27b & \textbf{0.22} & 0.15 & 0.70 & 0.13 \\
gpt-oss:20b (r=low) & 0.15 & 0.15 & 0.43 & 0.18 \\
gpt-oss:20b (r=medium) & \textbf{0.22} & \textbf{0.19} & 0.57 & \textbf{0.21} \\
qwen3:8b & 0.14 & 0.09 & 0.62 & 0.16 \\
llama3.1:8b & 0.14 & 0.13 & 0.65 & 0.17 \\
qwen2.5:3b & 0.14 & 0.12 & 0.43 & 0.17 \\
\bottomrule
\end{tabular}%
}
\end{center}
\appendixtablecaption{tab:wfc_attribute_breakdown_cont}{Continuation of Table~\ref{tab:wfc_attribute_breakdown}, reporting body, acidity, alcohol, and sugar accuracy for the WFC task. Best values per group are in bold.}

\end{document}